\documentclass[]{article}
\usepackage{layouts}
\usepackage{cite}
\usepackage{amsmath,amssymb,amsfonts}
\usepackage{algorithmic}
\usepackage{graphicx}
\usepackage{algorithm,algorithmic}
\usepackage{hyperref}
\hypersetup{hidelinks=true}
\usepackage{textcomp}
\usepackage{xcolor}
\usepackage{todonotes}

\usepackage{graphicx,verbatim}
\usepackage{tabularx}
\usepackage{multirow}
\usepackage[table]{xcolor}  
\usepackage{booktabs}
\usepackage{amssymb}
\usepackage{amsmath}

\usepackage{graphicx}
\usepackage{caption}

\usepackage{xcolor}
\usepackage{pifont}

\usepackage{acro}
\DeclareAcronym{ct}{short=CT, long=Computed Tomography}
\DeclareAcronym{rrg}{short=RRG, long=Radiology Report Generation}
\DeclareAcronym{ehr}{short=EHR, long=electronic health record}
\DeclareAcronym{cnn}{short=CNN, long=convolutional neural network}
\DeclareAcronym{vit}{short=ViT, long=vision transformer}
\DeclareAcronym{vlm}{short=VLM, long=Vision-Language Model}
\DeclareAcronym{llm}{short=LLM, long=Large-Language Model}
\DeclareAcronym{ve}{short=VE, long=Vision Encoder}
\DeclareAcronym{mae}{short=MAE, long=masked autoencoder}
\DeclareAcronym{nlp}{short=NLP, long=Natural Language Processing}
\DeclareAcronym{mlp}{short=MLP, long=Multilayer Perceptron}
\DeclareAcronym{bleu}{short=BLEU, long=Bilingual Evaluation Understudy}
\DeclareAcronym{roi}{short=ROI, long=region of interest}
\DeclareAcronym{fov}{short=FOV, long=field of view}

\usepackage{adjustbox}
\newsavebox{\mytabbox}
\let\oldtabular\tabular
\let\endoldtabular\endtabular
\renewenvironment{tabular}[2][c]{%
  \begin{lrbox}{\mytabbox}%
  \oldtabular[#1]{#2}%
}{%
  \endoldtabular
  \end{lrbox}%
  \adjustbox{max width=\textwidth}{\usebox{\mytabbox}}%
}

\title{Resolution Meets Reduction:\\Efficient Visual Context for\\3D Radiology Report Generation}

\author{%
  Jonathan~Suprijadi$^{1,*}$,
  Raphael~Stock$^{1,*}$,
  Moritz~Langenberg$^{1,*}$,\\
  David~Zimmerer$^1$,
  Kim-Celine~Kahl$^1$,
  Stefan~Denner$^1$,
  Yannick~Kirchhoff$^1$,\\
  Karol~Gotkowski$^1$,
  Maximilian~Rokuss$^1$,
  Jeremias~Traub$^1$,
  Tassilo~Wald$^1$,\\
  Constantin~Ulrich$^1$,
  and~Klaus~Maier-Hein$^1$%
}
\date{} 

\begin{document}
\maketitle
\thispagestyle{empty}
{\let\thefootnote\relax\footnotetext{$^*$Contributed equally. Co-first authors may list as lead on CV.}}
{\let\thefootnote\relax\footnotetext{$^1$All authors are or were with the Division of Medical Image Computing, German Cancer Research Center (DKFZ), Heidelberg, Germany (e-mail: jonathan.suprijadi@dkfz-heidelberg.de, raphael.stock@dkfz-heidelberg.de, moritz.langenberg@dkfz-heidelberg.de).}}

\begin{abstract}
Vision-language models offer a promising path toward automating radiology report generation, but applying them to full 3D CT volumes poses substantial computational challenges.
Modern foundation vision encoders (VEs) can produce tens of thousands of vision tokens per scan, making the visual sequence passed to the large language model (LLM) a primary computational bottleneck.
Vision-to-language projectors can compress this sequence to reduce computation, but may discard clinically relevant detail; conversely, effective compression can accommodate higher-resolution inputs while keeping the downstream token count fixed.
How this vision-token budget should be allocated across input field of view, spatial resolution, and vision-to-language projection therefore remains an open design question.
We systematically evaluate four heterogeneous VEs (CNN- and ViT-based), five token-reducing projectors at up to 64x compression alongside a non-reducing MLP projector baseline, 
and five instruction-tuned LLMs (1.7B--4B) on two large-scale CT report datasets (CT-RATE and Merlin).
At matched LLM token budgets, anatomy-guided region of interest cropping is the most consistent strategy, improving clinical macro F1 in 19 of 20 settings by +3.7 points on average for the 3D ViT Primus encoder and +1.1 for the slice-based 2D ViT Curia encoder.
Increasing input resolution further is strongly projector-dependent: the PerceiverResampler, paired with higher-resolution Curia features, yields the strongest configuration in the resolution study on both datasets.
Our best configurations achieve state-of-the-art clinical macro F1 on the test sets, reaching 49.5 on CT-RATE and 49.0 on Merlin.
Code and models will be published upon publication.

\vspace{0.5em}
\end{abstract}

\acresetall

\section{Introduction}

\begin{figure}[t!]
    \centering    
    \includegraphics[width=1.0\columnwidth]{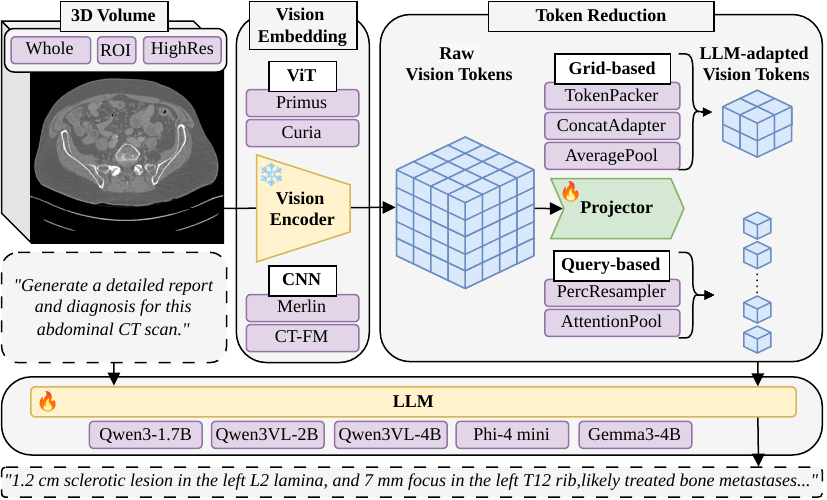}
    \caption{
    \textbf{Overview of our VLM pipeline for 3D radiology report generation}. Each 3D CT volume is first preprocessed according to one of three input regimes: whole-volume encoding, anatomy-guided ROI cropping, or higher-resolution encoding, before a frozen vision encoder extracts vision tokens. The projector compresses these tokens before they are passed to the LLM for report generation. 
    This work systematically investigates the design choices along this pipeline, including the input regime, vision encoder, projector, and LLM.
    }
    \label{fig:fig1}
\end{figure}

The diagnostic workload of radiologists has increased dramatically over the past two decades, and the interpretation of 3D \ac{ct} scans is among the most time-consuming tasks in clinical practice~\cite{forsberg2017radiologists}.
\acp{vlm} offer a promising solution by automating the drafting of diagnostic reports from imaging data~\cite{bannur2024maira,liu2026scaling}. 
Recent \ac{vlm} architectures consist of a pretrained \ac{ve} that produces vision tokens and a projector that maps them into the embedding space of a \acl{llm} (\acs{llm}), which then autoregressively generates the report~\cite{liu2023visual} (Fig.~\ref{fig:fig1}). 
The projector is typically a compression-free \ac{mlp} that forwards the entire token sequence to the \ac{llm}~\cite{wald2025comprehensive}.

Applying this paradigm to full 3D \ac{ct} volumes can produce extremely long visual contexts. Depending on the \ac{ve}, spatial resolution, and \ac{fov}, a single scan may yield tens of thousands of vision tokens, with higher-resolution inputs increasing this number further.
Common strategies for limiting the sequence length, such as downsampling to a fixed grid~\cite{bai2024m3d,hamamci2026ct-chat,blankemeier2026merlin}, \ac{roi} cropping~\cite{chen2025reg2rg,kalisch2025ctgraph}, or slice selection~\cite{sellergren2026medgemma}, reduce the amount of anatomy or spatial detail retained from the scan and may therefore suppress diagnostically relevant findings~\cite{sabottke2020effect}.
Preserving finer spatial detail, however, produces increasingly long and potentially redundant token sequences at substantial computational cost.
Recent work has addressed this challenge on the \ac{ve} side by learning more compact or structured volumetric representations, for example through learned volumetric tokenization~\cite{Hamamci2025BetterTF} or concept-specific visual aggregation~\cite{khlaut2026jolia}.
For pretrained \acp{ve}, the vision-to-language projector provides a complementary control point, aggregating the encoder representation into a compact context for report generation while regulating the downstream sequence length.
Yet despite this central role, projector choice in 3D \acp{vlm} is rarely studied as an independent design variable.
Current approaches span a single linear layer in Merlin~\cite{blankemeier2026merlin}, an MLP in COLIPRI~\cite{wald2025comprehensive}, 3D average pooling~\cite{bai2024m3d,du2024segvol}, attention pooling~\cite{hamamci2026ct-chat,Liu_2026_CVPR}, and Perceiver-style resampling in RadFM~\cite{wu2025radfm}, but their relative behavior across architectures and compression levels is not well characterized.
The closest prior study~\cite{baharoon2025exploring} explores only part of this space, but is constrained to 1,287 training cases~\cite{ji2022amos} and three projectors, one native to the evaluated \ac{ve} and thus plausibly favored.
We investigate these questions in two complementary experiments.

\subsubsection{Input Field of View and Resolution} 
We present, to the best of our knowledge, the first controlled study of input \ac{fov} and effective spatial resolution for 3D \ac{rrg} at a matched \ac{llm} token budget.
Holding the number of vision tokens forwarded to the \ac{llm} fixed separates the input regime from the sequence length the \ac{llm} attends to, and we compare whole-volume encoding, anatomy-guided \ac{roi} cropping, and two mechanisms for raising the effective resolution across two \acp{ve} and two datasets. \ac{roi} cropping is the most consistent intervention in the study, improving clinical macro F1 in 19 of 20 configurations, by $3.7$ on average for Primus~\cite{wald2025primus} and $1.1$ for Curia~\cite{dancette2025curia}, at the cost of a segmentation pass per volume.
Further increases in resolution do not improve performance on average and benefit only projectors whose aggregation mechanisms can effectively accommodate the stronger compression induced by larger inputs.

\subsubsection{Encoder, Projector, and Language Model} 
We report the broadest evaluation of encoder, projector, and language-model choices for 3D \ac{rrg} to date. Our study spans four medical imaging foundation \acp{ve} with \ac{cnn}- and \ac{vit}-based architectures, five token-reducing projectors evaluated at compression ratios up to $64\times$ alongside a token-preserving \ac{mlp} baseline, and five instruction-tuned \acp{llm} ranging from $1.7$B to $4$B parameters. Experiments are conducted on two large-scale CT image-report datasets with more than $25$k studies each, using three seeds per configuration and approximately $60{,}000$ A100 and $6{,}000$ H100 GPU-hours in total.
The study varies one factor at a time around a common development configuration, shows that the choice of the \ac{ve} has a larger impact than the choice of the \ac{llm}, and characterizes which projector families remain stable under aggressive compression. 
To facilitate clinical evaluation on Merlin, we release a lightweight ModernBERT~\cite{warner2024modernbert} classifier for 30 abnormality classes, trained on \ac{llm} extracted labels~\cite{openai2025gptoss120bgptoss20bmodel}, achieving 95.7 macro F1 on the held-out test split.

Together, these studies yield state-of-the-art clinical performance on both benchmarks, with a macro F1 of $49.5$ on CT-RATE against $48.0$~\cite{Liang2026BeyondTE} for the strongest comparable published system, and $49.0$ on Merlin against $23.8$~\cite{khlaut2026jolia}. 

\section{Methods}\label{sec:methods}

\subsection{Overall Architecture}

Our framework follows the standard three-stage LLaVA-style \ac{vlm} architecture consisting of a pretrained \ac{ve}, a vision-to-language projector, and a pretrained \ac{llm}. 
Given a 3D \ac{ct} volume, the \ac{ve} produces a sequence of $N$ visual feature tokens. The projector maps these features into the embedding space of the \ac{llm}, yielding $M$ projected vision tokens, where $M=N$ for token-preserving projectors and $M<N$ when token compression is applied. The projected tokens are prepended to the embeddings of a textual instruction and condition the autoregressive generation of the radiology report.
To isolate projector design from visual representation learning, the \ac{ve} remains frozen, while the projector and \ac{llm} are jointly trained. Unless otherwise specified, comparisons vary the projector architecture under a fixed \ac{llm}.

\subsection{Vision-Language Projectors}

For systematic comparison, we categorize the evaluated token-compressing projectors as grid-based or query-based and consider a token-preserving MLP as the baseline.

\subsubsection{Grid-Based Projectors} 

Grid-based projectors group neighboring vision tokens into local blocks according to the per-axis reduction factors $\mathbf{r}=(r_x,r_y,r_z)$, where a factor of $1$ indicates no reduction along the corresponding axis. Aggregating each block into a single representation reduces the token count from $N$ to $M=N/(r_xr_yr_z)$ while retaining the spatial organization of the encoder feature grid. The resulting tokens are subsequently mapped to the \ac{llm} embedding dimension using an MLP.

The \textit{AveragePool} projector~\cite{bai2024m3d} averages the tokens within each local block using pooling kernels and strides given by $\mathbf{r}=(r_x,r_y,r_z)$.

The \textit{ConcatAdapter}, following Qwen3-VL~\cite{bai2025qwen3}, concatenates the tokens within each local block along the feature dimension. This increases the input dimension of the subsequent projection by a factor of $r_xr_yr_z$ while avoiding information aggregation before projection.

\textit{TokenPacker}~\cite{li2025tokenpacker} constructs a low-resolution query grid by interpolating the encoder feature grid and uses the original high-resolution features as keys and values. Each output token thereby aggregates fine-grained information from its corresponding spatial region. For comparability with other projectors, we use only the final-layer encoder features rather than the multi-layer features employed in the original implementation.

\subsubsection{Query-Based Projectors}
Query-based projectors introduce a fixed set of~$L$ learnable tokens that aggregate information from all~$N$ vision tokens through cross-attention, decoupling the output length from the spatial resolution of the encoder features.

\textit{AttentionPool}~\cite{hamamci2026ct-chat} implements this aggregation using a single multi-head cross-attention layer.

The \textit{PerceiverResampler}~\cite{alayrac2022flamingo} extends this design by stacking multiple Perceiver layers, each composed of a cross-attention step followed by a 2-layer MLP with residual connections. Furthermore, the queries are included in the key-value sequence, enabling information exchange among queries.
While query-based projectors provide flexible and globally informed compression, they sacrifice the structured spatial topology maintained by grid-based projectors.

\subsubsection{Token-Preserving Baseline}
Following the LLaVA paradigm~\cite{liu2023visual}, an MLP independently maps each of the $N$ vision tokens into the \ac{llm} embedding space.
The original sequence length is retained, such that $M = N$, providing a compression-free baseline.

\subsubsection{Token-Budget Matching} \label{subsec:token_budget_matching}

The \acp{ve} considered in this work produce feature grids with substantially
different token counts depending on the encoder and preprocessing configuration
(Table~\ref{tab:input_preprocessing}). For each configuration, we evaluate one
or more reduction vectors $\mathbf{r}$. Each reduction vector defines a token
compression ratio of $N/M=r_xr_yr_z$ and a corresponding output token budget
of $M=N/(r_xr_yr_z)$. Different choices of $\mathbf{r}$ therefore correspond
to the evaluated compression levels.

To enable fair comparisons across projector families, the number of tokens
passed to the \ac{llm} is matched for a given \ac{ve}, preprocessing
configuration, and compression ratio. Grid-based projectors use the
corresponding reduction vector $\mathbf{r}$, whereas query-based projectors
use $L=M$ learnable queries. Both projector families therefore expose the
\ac{llm} to the same visual sequence length while employing different
aggregation mechanisms.

\subsection{Vision Encoders and Input Preprocessing}

\newcommand{\cmark}{\textcolor{green!50!black}{\ding{51}}}
\newcommand{\xmark}{\textcolor{red!70!black}{\ding{55}}}

\begin{table*}[t]
\centering
\caption{
\textbf{Dataset- and vision-encoder-specific input preprocessing.}
Token counts are reported before projection. ``Full'' denotes the full scan extent after whole-volume resizing. The VE input specifies the spatial input to a single encoder pass; Curia independently encodes 192 axial slices of $512^2$ pixels, whereas Primus-4B independently encodes four $192^3$-voxel blocks. For resampled inputs, Merlin and CT-FM use the voxel spacings of their original implementations, while Primus-4B uses isotropic $1$-mm spacing.
For resampled inputs:
\textsuperscript{a}$1\times1\times3$ mm;
\textsuperscript{b}$1.5\times1.5\times3$ mm;
\textsuperscript{c}$1\times1\times1$ mm.
}
\label{tab:input_preprocessing}

\setlength{\tabcolsep}{4pt}

\begin{tabular}{c l c c c c c c c c c}
\toprule
\multirow{2}{*}[-0.6ex]{\textbf{Setting}} &
\multirow{2}{*}[-0.6ex]{\textbf{VE}} &
\multicolumn{2}{c}{\textbf{Dataset}} &
\multirow{2}{*}[-0.6ex]{\textbf{Preproc.}} &
\multirow{2}{*}[-0.6ex]{%
    \shortstack{\textbf{VE input} \\ \textbf{per pass}}} &
\multirow{2}{*}[-0.6ex]{%
    \shortstack{\textbf{Block} \\ \textbf{layout}}} &
\multirow{2}{*}[-0.6ex]{%
    \shortstack{\textbf{Axial mean} \\ \textbf{pooling}}} &
\multirow{2}{*}[-0.6ex]{\textbf{FOV [mm]}} &
\multirow{2}{*}[-0.6ex]{%
    \shortstack{\textbf{VE output} \\ \textbf{grid}}} &
\multirow{2}{*}[-0.6ex]{%
    \shortstack{\textbf{Vision} \\ \textbf{tokens}}} \\
\cmidrule(lr){3-4}
& & \textbf{CT-RATE} & \textbf{Merlin} & & & & & & & \\
\midrule

\multirow{5}{*}[-1.6ex]{Baseline}
& Primus-1B
& \cmark & \cmark
& Resize
& $192^3$ voxels
& --
& --
& Full
& $24^3$
& 13,824 \\
\cmidrule(lr){2-11}

& Curia-8S
& \cmark & \cmark
& Resize
& $512^2$ pixels
& --
& $192 \to 8$
& Full
& $8 \times 32^2$
& 8,192 \\
\cmidrule(lr){2-11}

& CT-FM
& \cmark & \cmark
& Resample\textsuperscript{a}
& $192^3$ voxels
& --
& --
& $192 \times 192 \times 576$
& $12^3$
& 1,728 \\
\cmidrule(lr){2-11}

& Merlin
& \cmark & \cmark
& Resample\textsuperscript{b}
& $256^2 \times 160$ voxels
& --
& --
& $384 \times 384 \times 480$
& $8^2 \times 10$
& 640 \\

\midrule

\multirow{3}{*}[-0.9ex]{\shortstack{Higher-\\resolution}}
& \multirow{2}{*}{Primus-4B}
& \cmark & \xmark
& Resample\textsuperscript{c}
& \multirow{2}{*}{$192^3$ voxels}
& $2 \times 1 \times 2$
& --
& $384 \times 192 \times 384$
& \multirow{2}{*}{$4 \times 24^3$}
& \multirow{2}{*}{55,296} \\
&
& \xmark & \cmark
& Resize
&
& $2 \times 2 \times 1$
& --
& Full
&
& \\
\cmidrule(lr){2-11}

& Curia-64S
& \cmark & \cmark
& Resize
& $512^2$ pixels
& --
& $192 \to 64$
& Full
& $64 \times 32^2$
& 65,536 \\

\bottomrule
\end{tabular}
\end{table*}

\subsubsection{Region-of-Interest Cropping} \label{subsec:roi_cropping}
Before encoder-specific resizing or resampling, each volume is cropped to a target anatomical \ac{roi} automatically extracted using TotalSegmentator~\cite{wasserthal2023totalsegmentator}. The dataset-specific \ac{roi} definitions are provided in Section~\ref{subsec:datasets}. \ac{roi}-cropping removes non-body background and reduces the spatial extent allocated to irrelevant regions, allowing the target anatomy to be represented at a higher effective spatial resolution within the fixed encoder input and vision-token budget.

\subsubsection{Vision Encoders}

To assess whether projector recommendations are robust to backbone choice, we benchmark four medical imaging foundation \acp{ve}: Primus~\cite{wald2025primus}, Curia~\cite{dancette2025curia}, Merlin~\cite{blankemeier2026merlin}, and CT-FM~\cite{pai2025vision}. 
Together, they span \ac{cnn}- and transformer-based architectures, 2D and 3D inputs, and diverse pretraining objectives and scales.
Their architectural and pretraining characteristics are summarized in Table~\ref{tab:vision_encoders}, while the encoder-specific preprocessing and resulting feature-grid dimensions are reported in Table~\ref{tab:input_preprocessing}. Apart from the input-size adjustments described below, we retain the original preprocessing pipelines of Curia, CT-FM, and Merlin.

For the CNN-based \acp{ve} Merlin and CT-FM, we extract the bottleneck feature maps prior to the task-specific projection head to serve as volumetric token sequences. To ensure sufficient anatomical coverage, we increase the input sizes of CT-FM and Merlin from $128^2 \times 24$ to $192^3$ voxels and from $224^2 \times 160$ to $256^2 \times 160$ voxels, respectively, thereby enlarging their effective \acp{fov}. 
For CT-FM, we additionally explored resizing-based preprocessing and larger \acp{fov}, but neither improved performance on CT-RATE. We therefore retain the $192^3$ input throughout.

For Primus, we use the output tokens of the final transformer layer, which form a dense 3D token grid. The baseline configuration processes the volume as a single $192^3$ block and is denoted \mbox{\emph{Primus-1B}}, where ``B'' indicates the number of spatial blocks.

Curia is a 2D \ac{vit} that encodes axial slices independently. Each volume is resized to 192 axial slices with an in-plane resolution of $512^2$ pixels, matching the input size of the original Curia implementation. Axial mean pooling then aggregates the slice-level representations into eight axial feature slices, defining the baseline configuration \mbox{\emph{Curia-8S}}, where ``S'' denotes the number of retained axial feature slices after pooling. 
For Curia-8S, axial pooling already reduces the representation to only eight slices before projection, so projector compression is restricted to the transverse dimensions ($r_z=1$).

\begin{table}[t]
\centering
\caption{
\textbf{Comparison of pretrained vision encoders.}
}
\label{tab:vision_encoders}
\small
\setlength{\tabcolsep}{3.5pt}
\begin{tabular}{l c c c c c}
\toprule
\textbf{Model} & \textbf{Arch.} & \textbf{Param.} & \textbf{Dim.} &
\textbf{Obj.} & \textbf{Data} \\
\midrule
Primus & Primus-M & 150.0M & 3D &
MAE~\cite{he2022masked} & 2.2M vol. \\
Curia & ViT-B & 86.1M & 2D &
DINOv2~\cite{oquab2024dinov2} & 200M sl. \\
Merlin & ResNet152 & 121.9M & 3D &
CLIP~\cite{radford2021learning} & 25k CTs \\
CT-FM & SegResNet & 77.8M & 3D &
SimCLR~\cite{chen2020simple} & 148k CTs \\
\bottomrule
\end{tabular}
\end{table}

\subsection{Higher-Resolution Visual Representations}
\subsubsection{Primus-4B: Multi-Block Encoding}

To increase the effective spatial resolution relative to the single-block Primus-1B configuration, the cropped \ac{roi} is mapped to a four-block input grid by either whole-volume resizing or resampling to an isotropic voxel spacing of $1 \times 1 \times 1$~mm. This grid is partitioned into four contiguous, non-overlapping $192^3$-voxel blocks, each of which is encoded independently by the Primus \ac{ve}. The blockwise feature grids are reassembled according to their positions in the original tiling, thereby preserving the global spatial arrangement of the four blocks. We refer to this four-block approach as \mbox{\emph{Primus-4B}}.

\subsubsection{Curia-64S: Denser Axial Features}
While Primus-4B increases resolution through multi-block encoding, the 2D \ac{vit} architecture of Curia enables fine-grained resolution scaling along the vertical/longitudinal ($z$) axis.
Specifically, by reducing the kernel and stride of the axial mean pooling operation, we aggregate the 192 resampled 2D slices into 64 axial feature maps rather than the baseline 8.
We designate this configuration as \emph{Curia-64S}, which preserves richer anatomical detail along the axial dimension and results in an eightfold increase in the total number of vision tokens.

\subsection{Language Models}

We evaluate five instruction-tuned language models spanning 1.7B--4B parameters: Qwen3-1.7B~\cite{yang2025qwen3}, Qwen3-VL-2B and Qwen3-VL-4B~\cite{bai2025qwen3}, Phi-4-mini~\cite{abouelenin2025phi}, and Gemma-3-4B~\cite{kamath2025gemma}. Qwen3-1.7B and Phi-4-mini are text-only models, whereas Qwen3-VL-2B, Qwen3-VL-4B, and Gemma-3-4B are initialized from multimodally pretrained vision-language models. For the multimodal models, we replace the native vision pathway with our frozen vision encoder and vision-language projector. For Qwen3-VL, we leverage its multimodal rotary positional embeddings (M-RoPE)~\cite{bai2025qwen3} by mapping the axial dimension to the temporal axis for grid-based adapters. For query-based adapters, whose outputs are spatially unstructured, we use the text formulation, under which M-RoPE reduces to standard one-dimensional RoPE.

\section{Experiments}\label{sec:experiments}

\subsection{Datasets}\label{subsec:datasets}

As our development dataset, we use \emph{CT-RATE}~\cite{hamamci2026ct-chat}, comprising 50,188 non-contrast chest CT volumes (25,692 CT-report pairs from 21,304 patients), with multiple volumes per study reconstructed using different kernels.
Reports contain a free-text \textit{Findings} section and an optional \textit{Impression} summary.
We exclude 829 scans with a TotalSegmentator-estimated lung volume below 1\,L~\cite{wasserthal2023totalsegmentator} and two volumes whose associated report lacks a \textit{Findings} section.
Because no official test split is provided, we use the official validation set for testing and create a patient-disjoint validation split from the training data.
The resulting splits contain 22,590, 1,530, and 1,564 reports for training, validation, and testing, respectively.

For CT-RATE, the \ac{roi} is the axis-aligned bounding box enclosing the union of the five TotalSegmentator lung-lobe masks, expanded by 4\% of the side length along each axis.
This region covers the pulmonary and cardiovascular abnormalities targeted by the downstream evaluation while substantially reducing the input dimensions.

\begin{figure}[t]
    \centering
    \includegraphics[width=\columnwidth]{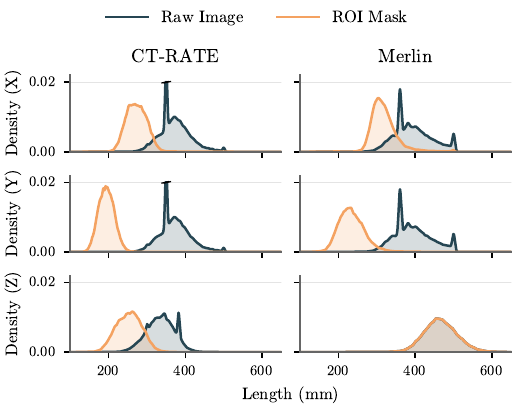}
    \caption{
        \textbf{Image dimension distributions before and after ROI cropping.} 
        Distributions along the X, Y, and Z axes are shown for the raw scans (dark blue) and cropped \acp{roi} (orange). CT-RATE uses a lung \ac{roi}, while Merlin uses a foreground \ac{roi}.
    }
    \label{fig:roi_vs_no_roi_dist}
\end{figure}

For cross-dataset evaluation, we use \emph{Merlin}~\cite{blankemeier2026merlin}, which contains 25,494 abdominal CT-report pairs.
We retain the official test set, remove eight duplicate training entries, and reserve 1,000 volumes for validation, yielding 19,361, 1,000, and 5,125 training, validation, and test cases, respectively.
Merlin differs substantially from CT-RATE in anatomical region, report structure, and clinical findings.
Because its evaluation spans most abdominal organs, as described in Section~\ref{subsec:evaluation_and_metrics}, we define a whole-body \ac{roi} from the union of all TotalSegmentator foreground classes.
After removing small disconnected components, we compute the foreground bounding box and expand it by 4\% along each axis.
The resulting shifts in image dimension distributions for both datasets are illustrated in Figure~\ref{fig:roi_vs_no_roi_dist}.

\subsection{Evaluation Metrics} \label{subsec:evaluation_and_metrics}

We use clinical macro F1 as the primary evaluation metric throughout all model-selection, hyperparameter-tuning, and ablation experiments. Lexical metrics are reported only for the final comparison of our best-performing configurations with existing methods and do not inform any development decisions.
Commonly used lexical metrics primarily measure surface-level similarity to reference reports and do not account for the clinical relevance or correctness of the generated content, making them unreliable indicators of medical report quality~\cite{li2025reevalmed,ostmeier2024green}.

\subsubsection{Clinical Evaluation}

For CT-RATE, we use the fine-tuned RadBERT classifier provided by Hamamci et al.~\cite{yan2022radbert,hamamci2026ct-chat}. The classifier predicts 18 pulmonary and cardiovascular abnormalities from free-text radiology reports. We apply it to both the generated and reference reports and compute the macro F1 score across all 18 classes.

Because no established compute-efficient clinical evaluation model is available for Merlin, we fine-tune ModernBERT~\cite{warner2024modernbert} (chosen over RadBERT~\cite{yan2022radbert} for its longer context, as ${\sim}45\%$ of Merlin reports exceed 512 tokens) as a multi-label classifier for the 30 abnormality classes of Blankemeier et al.~\cite{blankemeier2026merlin}, using labels generated by GPT-OSS-120B~\cite{openai2025gptoss120bgptoss20bmodel} instead of the original regular-expression-based extraction, inspired by~\cite{stock2026evaluating}. Trained on 20{,}361 reports (official training and validation splits), the classifier achieves a macro F1 of 0.957 against the GPT-OSS labels on the held-out official test split (5{,}125 reports). At evaluation time we apply it to the generated reports and compare against the GPT-OSS labels of the reference reports, reporting the macro F1 score across all 30 classes.

\subsubsection{Lexical Evaluation}

We additionally report BLEU-1 through BLEU-4~\cite{papineni2002bleu}, METEOR~\cite{banerjee2005meteor}, and ROUGE-L~\cite{lin2004rouge} using the implementations provided by the \texttt{pycocoevalcap} library~\cite{chen2015microsoft,pycocoevalcap}.

\subsection{Training Setup}

All models are trained using a next-token prediction objective with teacher forcing over the report text, conditioned on the vision tokens produced by the frozen \ac{ve} and trainable projector.
Unless stated otherwise, we use the instruction-tuned Qwen3-VL-2B~\cite{bai2025qwen3} as the shared \ac{llm} backbone across all experiments.
The projector is fully fine-tuned, while the \ac{llm} is adapted using LoRA~\cite{hu2022lora} with $r{=}32$, $\alpha{=}64$, and dropout$\,{=}\,0.1$.
Optimization uses AdamW~\cite{loshchilov2017decoupled} with $\beta_1{=}0.9$, $\beta_2{=}0.999$, and no weight decay~\cite{alayrac2022flamingo,wald2025comprehensive}.
Models are trained with a batch size of 24 for 16 epochs on CT-RATE and 32 epochs on Merlin, reflecting Merlin's approximately twofold smaller training set.
We perform validation after every epoch on CT-RATE and every two epochs on Merlin.
For each validation run, reports are generated and evaluated using the clinical macro F1 score.
The checkpoint achieving the highest validation F1 is selected for subsequent evaluation.

Training was performed on four-GPU nodes equipped with NVIDIA A100 40\,GB or NVIDIA H100 80\,GB GPUs, with individual runs using between 4 and 12 GPUs.
The single longest training run was the Primus MLP baseline, requiring roughly 240 H100 GPU-hours.

We adopt a two-stage linear warm-up schedule inspired by Wald et al.~\cite{wald2025openmind}.
In the first phase, only the projector is optimized to establish initial vision--language alignment.
In the second phase, the LoRA parameters are unfrozen and jointly optimized with the projector using a separate linear warm-up. Each warm-up phase spans $6.25\%$ ($1/16$) of the total optimization steps. For the PerceiverResampler, the initial projector-only warm-up is extended to $25\%$ ($4/16$), as preliminary experiments indicated that the standard duration was insufficient. After the two warm-up phases, the learning rate follows a cosine decay over the remaining steps.

\subsection{Hyperparameter Selection}\label{subsec:hyperparameters}

To limit computational cost and ensure a consistent search procedure across projectors, we perform the full hyperparameter search on the CT-RATE validation set using Primus-1B at an $8\times$ token-compression ratio, corresponding to $\mathbf{r}=(2,2,2)$.
For each projector, we evaluate the projector/\ac{llm} learning-rate pairs
$(3{\times}10^{-3},3{\times}10^{-4})$,
$(1{\times}10^{-3},3{\times}10^{-4})$,
$(3{\times}10^{-4},3{\times}10^{-4})$,
$(3{\times}10^{-4},1{\times}10^{-4})$, and
$(1{\times}10^{-4},1{\times}10^{-4})$.

We compare projector depth by evaluating 2- and 4-layer projection MLPs for all projectors except the PerceiverResampler.
For the PerceiverResampler, depth instead denotes the number of Perceiver layers, and we evaluate depths of three and six.

For each token-reducing projector--depth combination, we retain the three learning-rate pairs achieving the highest validation macro F1 at $8\times$ compression and evaluate them at the remaining Primus compression ratios of $27\times$ and $64\times$.
We then select the best-performing learning-rate and depth combination for each projector and compression ratio.

For each remaining \ac{ve}, we transfer the learning rate and projector depth from the Primus configuration at the corresponding or nearest available compression ratio.
No additional hyperparameter tuning is performed for other \acp{ve} or on Merlin, maintaining a consistent budget across comparisons.

\subsection{Experiment Design}

We conduct experiments on the CT-RATE and Merlin datasets. Unless otherwise stated, reported results are averaged over three seeds and presented as mean $\pm$ standard deviation. 

\subsubsection{Resolution and ROI Cropping}

We evaluate three input regimes, denoted \emph{Whole}, \emph{ROI}, and \emph{High Res}. \emph{Whole} encodes the complete resized volume without cropping, whereas \emph{ROI} applies the dataset-specific anatomical cropping procedure described in Section~\ref{subsec:roi_cropping}. \emph{High Res.} combines ROI cropping with increased effective resolution through blockwise encoding with Primus-4B or reduced post-encoding axial mean pooling with Curia-64S. The corresponding configurations are summarized in Table~\ref{tab:resolution_regimes}. 

For AveragePool, Curia-64S \emph{High Res.} is equivalent to Curia-8S \emph{ROI}, as projector pooling reproduces the preceding axial mean pooling.

\begin{table}[t]
\centering
\caption{\textbf{Resolution and ROI-cropping configurations under a fixed vision-token budget.} 
$\mathbf{r}$ denotes the grid-based reduction vector; query-based projectors use $L=M$ queries.}
\label{tab:resolution_regimes}
\setlength{\tabcolsep}{3.5pt}
\begin{tabular}{llcccc}
\toprule
\textbf{\ac{ve}} &
\textbf{Regime} &
\textbf{Configuration} &
\textbf{ROI} &
$\boldsymbol{\mathbf{r}}$ &
$\boldsymbol{M}$ \\
\midrule
\multirow{4}{*}{Primus} & Whole & Primus-1B & \xmark & $(2,2,2)$ & \multirow{4}{*}{1,728} \\
& ROI & Primus-1B & \cmark & $(2,2,2)$ & \\
& High Res. (CT-RATE) & Primus-4B & \cmark & $(4,2,4)$ & \\
& High Res. (Merlin) & Primus-4B & \cmark & $(4,4,2)$ & \\
\midrule
\multirow{3}{*}{Curia} & Whole & Curia-8S & \xmark & $(2,2,1)$ & \multirow{3}{*}{2,048} \\
& ROI & Curia-8S & \cmark & $(2,2,1)$ & \\
& High Res. & Curia-64S & \cmark & $(2,2,8)$ & \\
\bottomrule
\end{tabular}
\end{table}

\subsubsection{Token Compression}

To characterize how projector performance varies with compression strength, we evaluate token-reducing projectors at multiple encoder-specific reduction levels on ROI-cropped inputs.
Spatial reduction is applied uniformly across all three axes for 3D encoders, but restricted to the two transverse axes for Curia-8S.
Because native token counts differ, evaluated compression ratios vary across backbones: $8\times$ to $64\times$ for Primus-1B, $4\times$ to $64\times$ for Curia-8S, $8\times$ and $27\times$ for CT-FM, and $8\times$ for Merlin. The token-preserving MLP is included for every encoder as an uncompressed reference with $M=N$.

\subsubsection{\ac{llm} Choice}
To quantify the influence of the \ac{llm} on downstream performance, we evaluate multiple \acp{llm} while holding the visual configuration fixed. For each \ac{ve}-dataset pair, we retain the best-performing configuration from the resolution and ROI-cropping study. In addition to Qwen3-VL-2B, which is used throughout model development, we evaluate Qwen3-1.7B, Qwen3-VL-4B, Phi-4-mini, and Gemma3-4B.

\section{Results}

\subsection{Impact of Resolution and ROI Cropping}
\label{subsec:fov}

\begin{table}[tb]
\centering
\caption{\textbf{Effect of input field-of-view (FOV) / resolution.} Validation-set, seed-averaged macro F1 scores [\%] for increasing resolution (Whole $\to$ ROI $\to$ High Res.), with standard deviation across seeds ($\pm$ std).
The number of vision tokens $M$ forwarded to the LLM is fixed across FOV / resolution settings for each \ac{ve} (cf. Tables \ref{tab:input_preprocessing} and \ref{tab:resolution_regimes}).
Best per dataset and \ac{ve} in \textbf{bold}; color coding is applied per dataset and \ac{ve}. ``\emph{= ROI}'': equivalent to the corresponding ROI setting.}
\label{tab:fov}
\begingroup\fontsize{9}{11}\selectfont
\resizebox{\columnwidth}{!}{%
\begin{tabular}{ll||ccc||ccc}
\toprule
 & & \multicolumn{3}{c}{\textbf{Primus}} & \multicolumn{3}{c}{\textbf{Curia}} \\
\cmidrule(lr){3-5} \cmidrule(lr){6-8}
\textbf{Type} & \textbf{Projector} & \textbf{Whole} & \textbf{ROI} & \textbf{High Res.} & \textbf{Whole} & \textbf{ROI} & \textbf{High Res.} \\
\midrule
& & \multicolumn{6}{c}{\textbf{CT-RATE}} \\
\midrule
\multirow{3}{*}{\shortstack{Grid-\\based}} & AveragePool & \cellcolor[HTML]{E0C566} 44.6 \tiny{$\pm$ 0.5} & \cellcolor[HTML]{66B366} \textbf{47.3} \tiny{$\pm$ 0.7} & \cellcolor[HTML]{D3C366} 44.9 \tiny{$\pm$ 1.5} & \cellcolor[HTML]{E0C566} 47.8 \tiny{$\pm$ 0.5} & \cellcolor[HTML]{95BA66} 49.0 \tiny{$\pm$ 0.1} & \cellcolor[HTML]{E0E0E0} \emph{= ROI} \\
 & ConcatAdapter & \cellcolor[HTML]{FFB466} 43.2 \tiny{$\pm$ 0.7} & \cellcolor[HTML]{66B366} \textbf{47.3} \tiny{$\pm$ 0.8} & \cellcolor[HTML]{FFC566} 43.8 \tiny{$\pm$ 0.4} & \cellcolor[HTML]{D4C366} 48.0 \tiny{$\pm$ 0.4} & \cellcolor[HTML]{D1C266} 48.0 \tiny{$\pm$ 0.8} & \cellcolor[HTML]{FFBA66} 46.9 \tiny{$\pm$ 1.5} \\
 & TokenPacker & \cellcolor[HTML]{FFB666} 43.3 \tiny{$\pm$ 0.5} & \cellcolor[HTML]{A6BC66} 45.9 \tiny{$\pm$ 0.2} & \cellcolor[HTML]{E3C566} 44.6 \tiny{$\pm$ 1.0} & \cellcolor[HTML]{F7C866} 47.4 \tiny{$\pm$ 1.2} & \cellcolor[HTML]{C0C066} 48.3 \tiny{$\pm$ 1.0} & \cellcolor[HTML]{B3BE66} 48.5 \tiny{$\pm$ 0.7} \\
\cmidrule(lr){1-8}
\multirow{3}{*}{\raisebox{2.5ex}{\shortstack{Query-\\based}}} & AttentionPool & \cellcolor[HTML]{FF6666} 40.6 \tiny{$\pm$ 1.3} & \cellcolor[HTML]{FFBF66} 43.6 \tiny{$\pm$ 1.3} & \cellcolor[HTML]{FF7266} 41.0 \tiny{$\pm$ 0.7} & \cellcolor[HTML]{FF6766} 44.8 \tiny{$\pm$ 0.3} & \cellcolor[HTML]{FFB666} 46.8 \tiny{$\pm$ 0.4} & \cellcolor[HTML]{FF6666} 44.7 \tiny{$\pm$ 0.2} \\
 & PercResampler & \cellcolor[HTML]{DAC466} 44.8 \tiny{$\pm$ 0.6} & \cellcolor[HTML]{8FB966} 46.4 \tiny{$\pm$ 0.5} & \cellcolor[HTML]{9DBB66} 46.1 \tiny{$\pm$ 0.9} & \cellcolor[HTML]{CDC266} 48.1 \tiny{$\pm$ 0.4} & \cellcolor[HTML]{A8BC66} 48.7 \tiny{$\pm$ 0.4} & \cellcolor[HTML]{66B366} \textbf{49.8} \tiny{$\pm$ 1.0} \\
\midrule
& & \multicolumn{6}{c}{\textbf{Merlin}} \\
\midrule
\multirow{3}{*}{\shortstack{Grid-\\based}} & AveragePool & \cellcolor[HTML]{F4C766} 40.6 \tiny{$\pm$ 0.8} & \cellcolor[HTML]{94B966} 45.8 \tiny{$\pm$ 0.7} & \cellcolor[HTML]{ADBD66} 44.4 \tiny{$\pm$ 1.6} & \cellcolor[HTML]{EBC666} 44.8 \tiny{$\pm$ 1.0} & \cellcolor[HTML]{E5C566} 45.0 \tiny{$\pm$ 0.3} &  \cellcolor[HTML]{E0E0E0} \emph{= ROI} \\
 & ConcatAdapter & \cellcolor[HTML]{FFB066} 37.8 \tiny{$\pm$ 1.3} & \cellcolor[HTML]{CEC266} 42.6 \tiny{$\pm$ 1.6} & \cellcolor[HTML]{FF7B66} 33.4 \tiny{$\pm$ 0.9} & \cellcolor[HTML]{FFBF66} 43.7 \tiny{$\pm$ 0.6} & \cellcolor[HTML]{E5C566} 45.0 \tiny{$\pm$ 2.0} & \cellcolor[HTML]{FFB766} 43.3 \tiny{$\pm$ 0.7} \\
 & TokenPacker & \cellcolor[HTML]{C7C166} 42.9 \tiny{$\pm$ 0.5} & \cellcolor[HTML]{90B966} 46.0 \tiny{$\pm$ 1.6} & \cellcolor[HTML]{66B366} \textbf{48.3} \tiny{$\pm$ 1.8} & \cellcolor[HTML]{FFAA66} 42.7 \tiny{$\pm$ 1.3} & \cellcolor[HTML]{F6C866} 44.5 \tiny{$\pm$ 0.9} & \cellcolor[HTML]{A8BC66} 46.9 \tiny{$\pm$ 0.9} \\
\cmidrule(lr){1-8}
\multirow{3}{*}{\raisebox{2.5ex}{\shortstack{Query-\\based}}} & AttentionPool & \cellcolor[HTML]{FF7566} 32.8 \tiny{$\pm$ 1.2} & \cellcolor[HTML]{FFA766} 37.1 \tiny{$\pm$ 1.2} & \cellcolor[HTML]{FF6666} 31.6 \tiny{$\pm$ 2.7} & \cellcolor[HTML]{FF6666} 39.4 \tiny{$\pm$ 1.8} & \cellcolor[HTML]{FF9466} 41.6 \tiny{$\pm$ 2.3} & \cellcolor[HTML]{FF9066} 41.4 \tiny{$\pm$ 1.5} \\
 & PercResampler & \cellcolor[HTML]{FFB966} 38.6 \tiny{$\pm$ 0.6} & \cellcolor[HTML]{ACBD66} 44.5 \tiny{$\pm$ 0.4} & \cellcolor[HTML]{D9C466} 42.0 \tiny{$\pm$ 2.3} & \cellcolor[HTML]{FFB866} 43.3 \tiny{$\pm$ 1.0} & \cellcolor[HTML]{F8C866} 44.4 \tiny{$\pm$ 0.7} & \cellcolor[HTML]{66B366} \textbf{49.0} \tiny{$\pm$ 1.7} \\
\bottomrule
\end{tabular}}
\endgroup
\end{table}

\begin{figure*}
    \centering
    \includegraphics[width=1\linewidth]{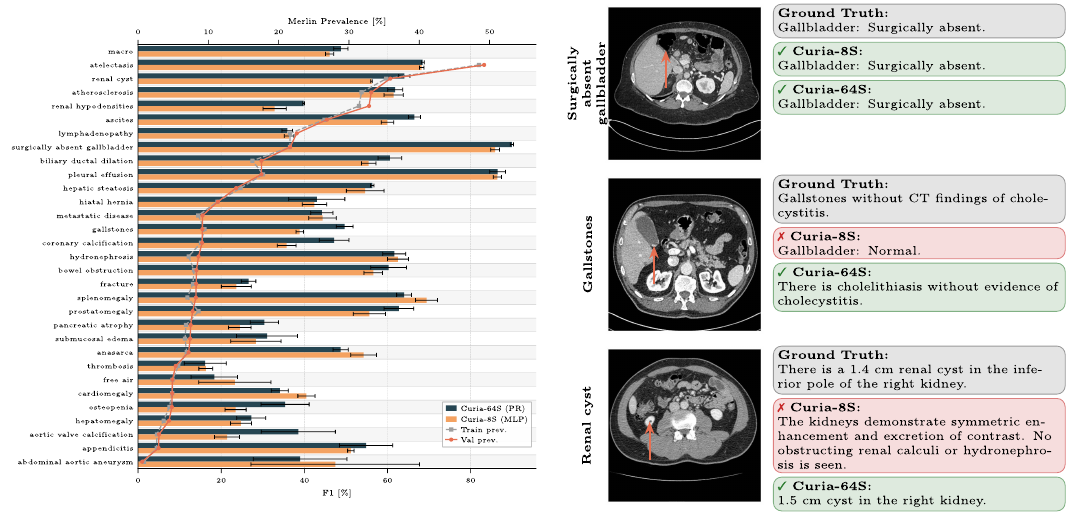}
    \caption{
    \textbf{Class-wise and qualitative results.}
    (left) Macro F1 (first row) and class-wise F1 scores [\%] for Curia-64S PerceiverResampler in \textcolor[HTML]{264653}{dark blue} (cf. Table \ref{tab:fov}) and Curia-8S MLP in \textcolor[HTML]{F4A261}{orange} (cf. Table \ref{tab:compression}) on the 30 classes from the Merlin dataset.
    All results are seed averaged including $\pm$ std.
    Classes are sorted by validation set prevalence.
    (right) Qualitative example of Curia-64S catching the small gallstone and renal cyst indicated by the \textcolor[HTML]{E76F51}{red arrows} while the lower resolution Curia-8S misses those small pathologies, highlighting the importance of resolution.
    }
    \label{fig:per-class_and_qualitative_res}
\end{figure*}

As summarized in Table~\ref{tab:fov}, cropping to the \ac{roi} improves F1 over the \emph{Whole} baseline in 19 of 20 configurations. 
The magnitude of this gain, however, differs systematically between the two encoders.
Averaged across all projectors and both datasets, the gain from \emph{Whole} to \emph{ROI} is smaller for Curia than for Primus ($+1.14$ vs.\ $+3.72$). This difference may reflect that Curia is already operating near native in-plane resolution, which limits the additional detail afforded by cropping, whereas Primus benefits more directly from the resulting increase in effective resolution.
For Primus, the average gain from \emph{Whole} to \emph{ROI} is larger on Merlin than on CT-RATE ($+4.63$ vs.\ $+2.82$).

On CT-RATE, the \emph{High Res.} regime does not yield a further improvement for Primus. Its best \emph{High Res.} run, obtained with the PerceiverResampler ($46.1$), remains below the best \emph{ROI} runs ($47.3$). We interpret this as a resolution ceiling effect: with a crop size of $192³$\,mm, the lung ROI already fits within a single Primus block at close to the \ac{ct}'s original resolution. For Curia, \emph{High Res.} is more beneficial, achieving the best overall CT-RATE result with the PerceiverResampler ($49.8$) and suggesting that reduced compression in $z$-slice aggregation preserves useful axial detail. This benefit is projector-dependent, though: TokenPacker is essentially flat ($48.3\to48.5$), whereas ConcatAdapter ($48.0\to46.9$) and AttentionPool ($46.8\to44.7$) degrade.

On Merlin, the \emph{High Res.} regime is more consistently beneficial, although its effect remains projector-dependent. For Primus, TokenPacker achieves the best result overall ($46.0\to48.3$), while the remaining projectors do not improve. For Curia, the PerceiverResampler gives the strongest result under \emph{High Res.} over \emph{ROI} ($44.4\to49.0$), and TokenPacker likewise improves ($44.5\to46.9$). 
By contrast, neither AveragePool, ConcatAdapter, nor AttentionPool benefits.

These results indicate that exploiting the additional input resolution depends on the projector's ability to aggregate a larger set of encoder tokens under stronger compression. Because the output token budget is held fixed, all \emph{High Res.} configurations use a compression ratio of $32\times$. The favorable results are concentrated in TokenPacker and the PerceiverResampler, suggesting that their aggregation mechanisms are better suited to preserve fine-grained information. 
In contrast, ConcatAdapter is particularly negatively affected because concatenating all tokens within each reduction block causes the input dimension and parameter count of the subsequent projection layer to scale with the compression ratio, an effect examined further in Section~\ref{subsec:compression}.

Finally, a per-class analysis (Figure~\ref{fig:per-class_and_qualitative_res}) indicates that the benefits of the \emph{High Res.} regime are especially evident for spatially small pathologies, for which the finer resolution available to the projector is most beneficial. 
At the fixed output token budget described above, \emph{High Res.} enables the projector to distill finer spatial detail into the same number of tokens. Specifically, renal cysts and gallstones on Merlin benefit significantly from the increased resolution.

\subsection{Impact of Token Compression}
\label{subsec:compression}

\begin{table*}[tb]
\centering
\caption{\textbf{Effect of token compression.} Validation-set, seed-averaged macro F1 scores [\%] in the \emph{ROI} input regime, with standard deviation across runs ($\pm$ std). Best per compression ratio and \ac{ve} in \textbf{bold}; color coding per metric, dataset, and \ac{ve}. \textit{MLP} rows use no token compression; each is shown once per encoder in the central compression column, in italic. 
}
\label{tab:compression}
\begingroup\fontsize{9}{11}\selectfont
\resizebox{\textwidth}{!}{%
\begin{tabular}{llll||ccc||ccc||cc||c}
\toprule
 &  &  &  & \multicolumn{3}{c}{\textbf{Primus}} & \multicolumn{3}{c}{\textbf{Curia}} & \multicolumn{2}{c}{\textbf{CT-FM}} & \multicolumn{1}{c}{\textbf{Merlin}} \\
\cmidrule(lr){5-7} \cmidrule(lr){8-10} \cmidrule(lr){11-12} \cmidrule(lr){13-13}
\textbf{Metric} & \textbf{Dataset} & \textbf{Type} & \textbf{Projector} & $8\times$ & $27\times$ & $64\times$ & $4\times$ & $16\times$ & $64\times$ & $8\times$ & $27\times$ & $8\times$ \\
\midrule
\multirow{12}{*}{Macro F1} & \multirow{6}{*}{CT-RATE} & MLP & \textit{(no compression)} &  & \textit{48.9} \tiny{$\pm$ 1.1} &  &  & \textit{48.5} \tiny{$\pm$ 0.8} &  &  \multicolumn{2}{c||}{\textit{35.6} \tiny{$\pm$ 0.6}} & \textit{32.0} \tiny{$\pm$ 0.9} \\
\cmidrule(lr){3-13}
 &  & \multirow{3}{*}{Grid-based} & AveragePool & \cellcolor[HTML]{66B366} \textbf{47.3} \tiny{$\pm$ 0.7} & \cellcolor[HTML]{A4BC66} \textbf{46.4} \tiny{$\pm$ 0.1} & \cellcolor[HTML]{FFC366} 45.0 \tiny{$\pm$ 0.6} & \cellcolor[HTML]{6BB366} \textbf{49.0} \tiny{$\pm$ 0.1} & \cellcolor[HTML]{C1C066} 47.1 \tiny{$\pm$ 2.0} & \cellcolor[HTML]{FFAD66} 44.7 \tiny{$\pm$ 0.7} & \cellcolor[HTML]{66B366} \textbf{35.6} \tiny{$\pm$ 0.7} & \cellcolor[HTML]{CAC166} 34.4 \tiny{$\pm$ 0.6} & \cellcolor[HTML]{C4C066} 31.9 \tiny{$\pm$ 0.3} \\
 &  &  & ConcatAdapter & \cellcolor[HTML]{66B366} \textbf{47.3} \tiny{$\pm$ 0.8} & \cellcolor[HTML]{FFB166} 44.7 \tiny{$\pm$ 1.1} & \cellcolor[HTML]{FF6666} 43.0 \tiny{$\pm$ 1.2} & \cellcolor[HTML]{97BA66} 48.0 \tiny{$\pm$ 0.8} & \cellcolor[HTML]{FF9866} 44.0 \tiny{$\pm$ 0.5} & \cellcolor[HTML]{FF6666} 42.2 \tiny{$\pm$ 0.3} & \cellcolor[HTML]{84B766} 35.2 \tiny{$\pm$ 0.6} & \cellcolor[HTML]{FDC966} 33.8 \tiny{$\pm$ 0.5} & \cellcolor[HTML]{66B366} \textbf{33.0} \tiny{$\pm$ 0.3} \\
 &  &  & TokenPacker & \cellcolor[HTML]{CAC166} 45.9 \tiny{$\pm$ 0.2} & \cellcolor[HTML]{C6C166} 46.0 \tiny{$\pm$ 0.7} & \cellcolor[HTML]{FFA766} 44.5 \tiny{$\pm$ 1.1} & \cellcolor[HTML]{8AB866} 48.3 \tiny{$\pm$ 1.0} & \cellcolor[HTML]{C6C166} 47.0 \tiny{$\pm$ 0.6} & \cellcolor[HTML]{FF9C66} 44.1 \tiny{$\pm$ 0.9} & \cellcolor[HTML]{FFC666} 33.7 \tiny{$\pm$ 2.1} & \cellcolor[HTML]{79B666} \textbf{35.4} \tiny{$\pm$ 0.9} & \cellcolor[HTML]{F0C766} 31.5 \tiny{$\pm$ 0.3} \\
\cmidrule(lr){3-13}
 &  & \multirow{2}{*}{Query-based} & AttentionPool & \cellcolor[HTML]{FF8066} 43.6 \tiny{$\pm$ 1.3} & \cellcolor[HTML]{FF7266} 43.3 \tiny{$\pm$ 1.6} & \cellcolor[HTML]{FF8A66} 43.8 \tiny{$\pm$ 0.8} & \cellcolor[HTML]{CDC266} 46.8 \tiny{$\pm$ 0.4} & \cellcolor[HTML]{FFA366} 44.4 \tiny{$\pm$ 1.3} & \cellcolor[HTML]{FFA666} 44.5 \tiny{$\pm$ 0.6} & \cellcolor[HTML]{C1C066} 34.5 \tiny{$\pm$ 0.8} & \cellcolor[HTML]{FF6666} 32.0 \tiny{$\pm$ 0.8} & \cellcolor[HTML]{FFBB66} 31.1 \tiny{$\pm$ 0.4} \\
 &  &  & PercResampler & \cellcolor[HTML]{A6BC66} 46.4 \tiny{$\pm$ 0.5} & \cellcolor[HTML]{C4C066} 46.0 \tiny{$\pm$ 0.3} & \cellcolor[HTML]{BCBF66} \textbf{46.1} \tiny{$\pm$ 0.3} & \cellcolor[HTML]{79B666} 48.7 \tiny{$\pm$ 0.4} & \cellcolor[HTML]{73B566} \textbf{48.8} \tiny{$\pm$ 0.7} & \cellcolor[HTML]{66B366} \textbf{49.2} \tiny{$\pm$ 0.4} & \cellcolor[HTML]{88B866} 35.2 \tiny{$\pm$ 0.3} & \cellcolor[HTML]{90B966} 35.1 \tiny{$\pm$ 0.3} & \cellcolor[HTML]{FF6666} 29.6 \tiny{$\pm$ 0.4} \\
\cmidrule(lr){2-13}
 & \multirow{6}{*}{Merlin} & MLP & \textit{(no compression)} &  & \textit{49.2} \tiny{$\pm$ 1.0} &  &  & \textit{46.3} \tiny{$\pm$ 1.0} &  &  \multicolumn{2}{c||}{\textit{28.2} \tiny{$\pm$ 1.5}} & \textit{33.2} \tiny{$\pm$ 0.9} \\
\cmidrule(lr){3-13}
 &  & \multirow{3}{*}{Grid-based} & AveragePool & \cellcolor[HTML]{72B566} 45.8 \tiny{$\pm$ 0.7} & \cellcolor[HTML]{CBC166} 41.4 \tiny{$\pm$ 1.2} & \cellcolor[HTML]{FFBC66} 38.0 \tiny{$\pm$ 1.5} & \cellcolor[HTML]{68B366} \textbf{45.0} \tiny{$\pm$ 0.3} & \cellcolor[HTML]{CDC266} 41.3 \tiny{$\pm$ 1.0} & \cellcolor[HTML]{FF8E66} 36.3 \tiny{$\pm$ 1.5} & \cellcolor[HTML]{66B366} \textbf{28.7} \tiny{$\pm$ 0.9} & \cellcolor[HTML]{D2C266} 26.0 \tiny{$\pm$ 1.1} & \cellcolor[HTML]{66B366} \textbf{33.4} \tiny{$\pm$ 0.9} \\
 &  &  & ConcatAdapter & \cellcolor[HTML]{B3BE66} 42.6 \tiny{$\pm$ 1.6} & \cellcolor[HTML]{FF9966} 35.3 \tiny{$\pm$ 0.5} & \cellcolor[HTML]{FF6666} 31.4 \tiny{$\pm$ 1.5} & \cellcolor[HTML]{68B366} 45.0 \tiny{$\pm$ 2.0} & \cellcolor[HTML]{FFB166} 38.2 \tiny{$\pm$ 0.3} & \cellcolor[HTML]{FF6666} 34.0 \tiny{$\pm$ 0.5} & \cellcolor[HTML]{B4BE66} 26.8 \tiny{$\pm$ 2.1} & \cellcolor[HTML]{FF8A66} 22.6 \tiny{$\pm$ 0.4} & \cellcolor[HTML]{A3BC66} 33.2 \tiny{$\pm$ 0.7} \\
 &  &  & TokenPacker & \cellcolor[HTML]{6DB466} \textbf{46.0} \tiny{$\pm$ 1.6} & \cellcolor[HTML]{66B366} \textbf{46.4} \tiny{$\pm$ 0.9} & \cellcolor[HTML]{AABD66} \textbf{43.0} \tiny{$\pm$ 0.6} & \cellcolor[HTML]{77B566} 44.5 \tiny{$\pm$ 0.9} & \cellcolor[HTML]{8CB866} 43.7 \tiny{$\pm$ 1.4} & \cellcolor[HTML]{FDC966} 39.6 \tiny{$\pm$ 0.4} & \cellcolor[HTML]{A1BB66} 27.2 \tiny{$\pm$ 1.9} & \cellcolor[HTML]{DAC466} 25.8 \tiny{$\pm$ 1.3} & \cellcolor[HTML]{CBC166} 33.1 \tiny{$\pm$ 0.4} \\
\cmidrule(lr){3-13}
 &  & \multirow{2}{*}{Query-based} & AttentionPool & \cellcolor[HTML]{FFB166} 37.1 \tiny{$\pm$ 1.2} & \cellcolor[HTML]{EFC766} 39.7 \tiny{$\pm$ 2.5} & \cellcolor[HTML]{E9C666} 40.0 \tiny{$\pm$ 1.3} & \cellcolor[HTML]{C6C166} 41.6 \tiny{$\pm$ 2.3} & \cellcolor[HTML]{F5C866} 39.9 \tiny{$\pm$ 1.6} & \cellcolor[HTML]{E6C566} 40.4 \tiny{$\pm$ 0.7} & \cellcolor[HTML]{FF6666} 21.2 \tiny{$\pm$ 1.2} & \cellcolor[HTML]{FF7966} 22.0 \tiny{$\pm$ 1.1} & \cellcolor[HTML]{FF6666} 32.4 \tiny{$\pm$ 0.5} \\
 &  &  & PercResampler & \cellcolor[HTML]{8CB866} 44.5 \tiny{$\pm$ 0.4} & \cellcolor[HTML]{A9BD66} 43.1 \tiny{$\pm$ 1.7} & \cellcolor[HTML]{B5BE66} 42.5 \tiny{$\pm$ 0.6} & \cellcolor[HTML]{79B666} 44.4 \tiny{$\pm$ 0.7} & \cellcolor[HTML]{66B366} \textbf{45.1} \tiny{$\pm$ 0.1} & \cellcolor[HTML]{A1BB66} \textbf{42.9} \tiny{$\pm$ 1.2} & \cellcolor[HTML]{B3BE66} 26.8 \tiny{$\pm$ 1.2} & \cellcolor[HTML]{CCC266} \textbf{26.2} \tiny{$\pm$ 0.6} & \cellcolor[HTML]{FFA266} 32.7 \tiny{$\pm$ 1.3} \\
\bottomrule
\end{tabular}}
\endgroup
\end{table*}

Table~\ref{tab:compression} reports performance across the evaluated compression ratios. 
For the \ac{vit}-based encoders, token-reducing projectors exceed the uncompressed \ac{mlp} reference only in a few cases.
For Primus, the \ac{mlp} achieves the best overall result on both datasets ($48.9$ on CT-RATE and $49.2$ on Merlin), and no token-reducing projector matches its performance. On CT-RATE, AveragePool and ConcatAdapter yield the strongest compressed results at $8\times$ ($47.3$), whereas the PerceiverResampler performs best at $64\times$ ($46.1$). On Merlin, TokenPacker is the strongest compressed projector at every evaluated ratio, achieving $46.0$, $46.4$, and $43.0$ at $8\times$, $27\times$, and $64\times$, respectively.

Curia exhibits a different pattern. The PerceiverResampler is the strongest projector at higher compression ratios and surpasses the uncompressed \ac{mlp} reference on CT-RATE at $64\times$ ($49.2$ vs.\ $48.5$). On Merlin, it likewise performs best among the token-reducing projectors at $16\times$ and $64\times$ ($45.1$ and $42.9$, respectively), but remains below the \ac{mlp} reference ($46.3$).
These results are consistent with the analysis in Section~\ref{subsec:fov}, where TokenPacker and the PerceiverResampler likewise proved most effective at preserving useful information under strong compression in the \emph{High Res.} regime.

The clearest failure mode under aggressive compression is observed with the ConcatAdapter, whose input dimension and parameter count grow with the compression ratio, as previously noted in Section~\ref{subsec:fov}.
For Primus, the ConcatAdapter increases from $39.4$ million parameters at $8\times$ compression to $369.2$ million at $64\times$, corresponding to a $9.4\times$ increase despite capping the first hidden dimension.
This substantial growth increases the projector's optimization burden and may contribute to its sensitivity to aggressive compression.
In additional experiments, an extended projector-only warm-up did not improve performance, indicating that insufficient warm-up alone does not explain the degradation. On Merlin, ConcatAdapter performance drops by more than $10$ F1 points over the same compression range for both Primus and Curia.
AveragePool exhibits a similar decline, likely because averaging over large token groups removes fine-grained spatial information that cannot be recovered by the subsequent projection. 
At the lowest compression ratio, however, it remains highly competitive and achieves some of the strongest results overall.

The \ac{cnn}-based encoders yield substantially lower absolute scores than Primus and Curia on both datasets. This may reflect the stronger spatial downsampling already performed by the \ac{cnn} backbones, which produces shorter token sequences and may leave less fine-grained information for the projector to preserve.
At $8\times$ compression, AveragePool is strongest in three of the four encoder--dataset combinations and matches the corresponding uncompressed \ac{mlp} reference in all four.
The exception is the Merlin encoder on CT-RATE, where ConcatAdapter performs best and surpasses the \ac{mlp} ($33.0$ vs.\ $32.0$).
At $27\times$ with CT-FM, TokenPacker is strongest on CT-RATE ($35.4$), whereas the PerceiverResampler is strongest on Merlin ($26.2$), consistent with their favorable behavior under stronger compression observed above.

Across settings with multiple compression ratios, AttentionPool varies comparatively little as compression increases, a pattern also observed for the PerceiverResampler. 
This stability may represent an advantage of query-based projectors: learned queries aggregate information adaptively from the full encoder feature set rather than summarizing increasingly large fixed spatial blocks. 
AttentionPool nevertheless underperforms compared to the other token-reducing projectors, suggesting that a single cross-attention layer may be insufficient to extract and refine the relevant visual information. The stronger results of the PerceiverResampler indicate that deeper, iterative query-based aggregation is needed to combine robustness to compression with high predictive performance.

\begin{table}[tb]
\centering
\caption{
\textbf{Effect of the language model.}
For each dataset, the vision path is fixed to the best configuration from Table~\ref{tab:fov}, indicated in parentheses after each vision encoder (HR: \emph{High Res.}; AP: AveragePool; TP: TokenPacker; PR: PerceiverResampler).
Values report validation-set seed-averaged macro F1 [\%] with standard deviation across seeds ($\pm$ std).
The best \ac{llm} per row is shown in \textbf{bold}; cells are color-coded within rows.}
\label{tab:best_per_encoder_llm}
\begingroup\fontsize{9}{11}\selectfont
\resizebox{\columnwidth}{!}{%
\begin{tabular}{l||ccccc}
\toprule
\multirow{2}{3cm}[-3.6ex]{\centering\textbf{VE,\\ (FOV/Res., projector)}} &
\multicolumn{5}{c}{\textbf{Language Model}} \\
\cmidrule(lr){2-6}
& \textbf{Qwen3} & \textbf{Qwen3VL} & \textbf{Phi-4} & \textbf{Gemma3} & \textbf{Qwen3VL} \\
& \textbf{1.7B} & \textbf{2B} & \textbf{mini} & \textbf{4B} & \textbf{4B} \\
\midrule

& \multicolumn{5}{c}{\textbf{CT-RATE}} \\
\midrule
Primus (\emph{ROI}, AP) &
\cellcolor[HTML]{ADBD66} 46.5 \tiny{$\pm$ 0.2} &
\cellcolor[HTML]{66B366} \textbf{47.3} \tiny{$\pm$ 0.7} &
\cellcolor[HTML]{FF6666} 43.8 \tiny{$\pm$ 0.9} &
\cellcolor[HTML]{FFB266} 45.1 \tiny{$\pm$ 0.7} &
\cellcolor[HTML]{FF7766} 44.1 \tiny{$\pm$ 0.7} \\
Curia (\emph{HR}, PR) &
\cellcolor[HTML]{E4C566} 49.2 \tiny{$\pm$ 0.5} &
\cellcolor[HTML]{88B866} 49.8 \tiny{$\pm$ 1.0} &
\cellcolor[HTML]{FF6666} 47.9 \tiny{$\pm$ 0.4} &
\cellcolor[HTML]{66B366} \textbf{50.1} \tiny{$\pm$ 0.6} &
\cellcolor[HTML]{E5C566} 49.2 \tiny{$\pm$ 1.5} \\

\midrule
& \multicolumn{5}{c}{\textbf{Merlin}} \\
\midrule
Primus (\emph{HR}, TP) &
\cellcolor[HTML]{9DBB66} 44.2 \tiny{$\pm$ 4.1} &
\cellcolor[HTML]{72B566} 48.3 \tiny{$\pm$ 1.8} &
\cellcolor[HTML]{8CB866} 45.8 \tiny{$\pm$ 1.8} &
\cellcolor[HTML]{FF6666} 20.5 \tiny{$\pm$ 0.8} &
\cellcolor[HTML]{66B366} \textbf{49.5} \tiny{$\pm$ 0.3} \\
Curia (\emph{HR}, PR) &
\cellcolor[HTML]{FF7366} 46.5 \tiny{$\pm$ 1.9} &
\cellcolor[HTML]{66B366} \textbf{49.0} \tiny{$\pm$ 1.7} &
\cellcolor[HTML]{FF7D66} 46.6 \tiny{$\pm$ 1.1} &
\cellcolor[HTML]{FF6666} 46.3 \tiny{$\pm$ 0.6} &
\cellcolor[HTML]{D1C266} 48.1 \tiny{$\pm$ 1.6} \\

\bottomrule
\end{tabular}
}
\endgroup
\end{table}

\begin{table*}[tb]
\centering
\caption{
\textbf{Comparison with the literature.}
We report seed-averaged test-set results for two configurations: Curia-64S with the PerceiverResampler at $32\times$ compression, using the best language model for each dataset, and Primus-1B with an MLP.
For CT-RATE baselines, we report published metrics.
For the Merlin and Jolia baselines, we apply our evaluation framework to their official test-set reports.
Our Merlin configurations are evaluated on the \emph{Findings} section and full report.
Values are in [\%]; the best per column and dataset is shown in \textbf{bold}.
Missing entries indicate metrics not reported in the respective publication.
}
\label{tab:literature}
\begingroup\fontsize{9}{11}\selectfont
\resizebox{\textwidth}{!}{%
\begin{tabular}{ll||ccccc|cccccc}
\toprule
 &  & \multicolumn{5}{c|}{\textbf{Clinical}} & \multicolumn{6}{c}{\textbf{Lexical}} \\
\cmidrule(lr){3-7}\cmidrule(lr){8-13}
\textbf{Dataset} & \textbf{Model} & \textbf{F1} & \textbf{Bal-Acc} & \textbf{Acc} & \textbf{Recall} & \textbf{Prec.} & \textbf{BLEU-1} & \textbf{BLEU-2} & \textbf{BLEU-3} & \textbf{BLEU-4} & \textbf{METEOR} & \textbf{ROUGE-L} \\
\midrule
\multirow{8}{*}{CT-RATE} & \textbf{Ours} \small(Curia-64S + PercResampler, Gemma-3-4B) & \textbf{49.5} & \textbf{68.4} & 83.1 & 46.7 & 52.8 & 46.0 & 33.3 & 25.4 & 20.3 & 23.5 & 28.0 \\
 & \textbf{Ours} \small(Primus-1B + MLP, Qwen3-VL-2B) & 49.0 & 68.0 & \textbf{84.3} & 43.8 & \textbf{57.0} & 42.9 & 31.6 & 24.7 & 20.2 & 22.8 & 31.0 \\
 & AdaRAG-CT (Llama-3.1)~\cite{Liang2026BeyondTE} & 48.0 & -- & -- & 52.0 & 50.2 & 49.6 & -- & -- & 24.2 & \textbf{24.6} & \textbf{35.4} \\
 & COLIPRI-CRM~\cite{wald2025comprehensive} & 44.9 & -- & -- & -- & -- & -- & -- & -- & -- & -- & -- \\
 & Ker-VLJEPA-3B (Phase 4)~\cite{Bumgardner2026CurriculumDriven3C} & 42.9 & -- & -- & \textbf{52.4} & 38.9 & -- & -- & -- & -- & -- & -- \\
 & CT-Agent~\cite{Mao2025CTAgentAM} & 42.0 & -- & -- & 47.7 & 42.3 & \textbf{50.2} & \textbf{37.4} & 29.0 & 23.1 & -- & -- \\
 & U-VLM~\cite{Shi2026UVLMHV} & 41.4 & -- & -- & 42.9 & 49.1 & 47.4 & 36.7 & \textbf{30.0} & \textbf{25.6} & -- & -- \\
 & BTB3D~\cite{Hamamci2025BetterTF} & 25.8 & -- & -- & 26.0 & 26.0 & 43.9 & 32.0 & 24.8 & 21.3 & 22.3 & -- \\
 
\cmidrule(lr){1-13}

\multirow{4}{*}{\shortstack[l]{Merlin\\(Findings)}} & \textbf{Ours} \small(Curia-64S + PercResampler, Qwen3-VL-2B) & 46.8 & 68.4 & \textbf{89.5} & 42.1 & 54.7 & 35.1 & \textbf{24.4} & \textbf{18.1} & \textbf{14.3} & \textbf{18.3} & \textbf{31.6} \\
 & \textbf{Ours} \small(Primus-1B + MLP, Qwen3-VL-2B) & \textbf{49.0} & \textbf{69.5} & \textbf{89.5} & \textbf{44.4} & \textbf{56.3} & \textbf{35.3} & 24.0 & 17.5 & 13.5 & 18.2 & 30.1 \\
 & Jolia~\cite{khlaut2026jolia} & 23.8 & 58.4 & 87.5 & 21.4 & 39.5 & 13.7 & 10.0 & 7.8 & 6.4 & 14.1 & 29.3 \\
 & Merlin~\cite{blankemeier2026merlin} & 9.7 & 53.0 & 87.5 & 9.6 & 21.6 & 10.9 & 7.8 & 5.8 & 4.6 & 12.9 & 25.8 \\
 
\cmidrule(lr){1-13}

\multirow{2}{*}{\raisebox{-0.5ex}{\shortstack[l]{Merlin\\(Full report)}}} & \textbf{Ours} \small(Curia-64S + PercResampler, Qwen3-VL-2B) & 48.3 & 69.5 & \textbf{89.6} & 44.4 & 54.2 & 33.0 & \textbf{22.3} & \textbf{16.4} & \textbf{12.8} & \textbf{17.0} & \textbf{29.0} \\
 & \textbf{Ours} \small(Primus-1B + MLP, Qwen3-VL-2B) & \textbf{50.0} & \textbf{70.3} & \textbf{89.6} & \textbf{46.3} & \textbf{55.3} & \textbf{33.6} & 22.2 & 15.9 & 12.1 & 16.9 & 27.6 \\

\bottomrule
\end{tabular}}
\endgroup
\end{table*}

\subsection{Impact of \ac{llm} Choice} \label{subsec:llm}
Results are reported in Table~\ref{tab:best_per_encoder_llm}, with the vision path fixed to its best configuration from Section~\ref{subsec:fov}. Excluding one failed run discussed below, performance varies by only $2.2$--$5.3$ F1 across \acp{llm} for a fixed dataset and \ac{ve}, substantially less than the more than $10$-point gap between the \ac{vit}- and \ac{cnn}-based encoders in Table~\ref{tab:compression}. Within the evaluated design space, the \ac{ve} appears to be the more influential component.

Qwen3-1.7B and Qwen3-VL-2B provide a near-controlled comparison of language-only and vision--language pretraining, as their language backbones are parameter-identical and the original vision tower of Qwen3-VL-2B is replaced by our medical \ac{ve}. Qwen3-VL-2B nevertheless performs better in all four settings, by $0.6$--$0.8$ on CT-RATE and $2.5$--$4.1$ on Merlin. This suggests that vision--language pretraining improves the language backbone's ability to consume visual embeddings even when the original \ac{ve} is discarded. Parameter count alone is not predictive: the larger Phi-4-mini does not outperform Qwen3-VL-2B in any setting, while Gemma-3-4B and Qwen3-VL-4B each improve upon it in only one of four settings.

These comparisons remain subject to transferred hyperparameters. Qwen3-VL-2B served as the development \ac{llm}, and repeating the learning-rate search for every backbone was computationally infeasible. 
The remaining results may therefore underestimate the performance attainable with model-specific tuning. This issue is most apparent for Gemma-3-4B with Primus on Merlin, which reproducibly collapses to $20.5$ F1 despite performing strongly elsewhere, highlighting the need to retune the language-side optimization when exchanging the \ac{llm}.

\subsection{Comparison to State-of-the-art}
\label{subsec:sota}

Table~\ref{tab:literature} compares our best configurations against published results on the held-out test splits.
We report the strongest token-reducing configuration, Curia-64S with the PerceiverResampler at $32\times$ compression, alongside the strongest uncompressed baseline, Primus-1B with an MLP. Gemma-3-4B is used for the token-reducing configuration on CT-RATE, while Qwen3-VL-2B is used otherwise.

On CT-RATE, Curia-64S with the PerceiverResampler achieves the highest F1 among the compared systems ($49.5$), exceeding AdaRAG-CT ($48.0$). The Primus MLP configuration attains the highest precision ($57.0$), while the recall of the Curia configuration ($46.7$) remains below that of Ker-VLJEPA-3B ($52.4$) and AdaRAG-CT ($52.0$). These results are obtained through end-to-end report generation, without the retrieval augmentation used by AdaRAG-CT~\cite{Liang2026BeyondTE} or the multi-stage agentic refinement used by CT-Agent~\cite{Mao2025CTAgentAM}. Such mechanisms are orthogonal to the projector design studied here and could plausibly be combined with our recipe for potential further gains.
CT-Agent and U-VLM obtain the highest BLEU scores, while AdaRAG-CT leads METEOR and ROUGE-L ($24.6$ and $35.4$). Nevertheless, CT-Agent and U-VLM trail our Curia configuration by $7.5$ and $8.1$ F1, respectively. This divergence is consistent with the limitations discussed in Section~\ref{subsec:evaluation_and_metrics}: n-gram overlap rewards similarity to the reference report style, but does not directly measure whether clinically relevant findings are reported correctly.

On the Merlin findings-only evaluation, our models ($46.8$, $49.0$) outperform Jolia and Merlin ($23.8$, $9.7$) across all reported metrics.
Evaluating the full report further raises F1 for Curia and Primus ($46.8\to48.3$ and $49.0\to50.0$) and improves recall, although precision decreases slightly, suggesting that clinically relevant labels are also conveyed outside the organ-wise \emph{Findings} section. Primus with the MLP achieves stronger clinical performance, whereas Curia with the PerceiverResampler performs better on every lexical metric except BLEU-1.
\section{Discussion, Limitations \& Conclusion}

This work asks where the vision-token budget is best spent.
Cropping to the target anatomy raises the effective resolution at an unchanged token budget and improves clinical macro F1 in 19 of 20 \ac{ve}-projector-dataset cells, by $3.7$ F1 for Primus and $1.1$ for Curia on average, which makes it the most consistent intervention in this study (Table~\ref{tab:fov}).
Resolution beyond that point is not free. 
When the emitted token budget is held fixed, a larger input must be compensated by a higher compression ratio and results in an average performance decrease of $1.2$ F1 on CT-RATE and $1.1$ F1 on Merlin. Only TokenPacker and the PerceiverResampler benefit from the higher-resolution regime, with their aggregation mechanisms better preserving useful information under stronger compression. Together, they yield the best results on 3 out of 4 \ac{ve}-dataset configurations.
We employ a lung-crop on CT-RATE specifically to ablate the effects of resolution in \ac{rrg}: a lung-crop suits the closed-set clinical question we evaluate, but it also removes anatomy the reference reports describe, which can encourage statements about structures the input no longer contains.

Within the \ac{vlm} design, the \ac{ve} is the dominant axis.
Averaged over all projectors and compression ratios, the spread between the best and the weakest \ac{ve} is $15.0$ F1 on CT-RATE and $16.1$ on Merlin. 
The Merlin \ac{ve} performs worse than CT-FM on CT-RATE but better on its in-domain dataset, underscoring the role of both architecture and pre-training data.
Across our four \acp{ve}, architecture, pretraining objective, pretraining scale and native token count all covary: both \ac{vit} encoders are pretrained by masking or self-distillation on $2.2$M volumes and $200$M slices and emit $8$k--$14$k tokens, whereas both \ac{cnn} encoders are pretrained contrastively on $25$k--$148$k volumes and emit $640$--$1{,}728$. 
Separating these factors would require pretraining matched architectures on a common corpus, which was out of scope here on compute grounds; we consider it a worthwhile target for future investigations.

Projector choice has a smaller effect overall: averaged over \acp{ve} and compression ratios, the difference between the most and least effective projector is $2.35$ F1 on CT-RATE and $4.47$ on Merlin.
One key finding is that most token-reducing projectors do not match the token-preserving \ac{mlp}. 
We therefore recommend using it as the reference when evaluating projectors.
Where compression is required, the choice is \ac{ve}-dependent: The PerceiverResampler exceeds its \ac{mlp} baseline ($49.2$ vs.\ $48.5$ at $64\times$ for Curia) while TokenPacker is the strongest compression projector for Primus on Merlin. 
When excluding outliers, the \ac{llm} choice has an impact similar to that of the projector.
Scores within the same \ac{ve} configuration span $2.2$--$5.3$ F1 points, but the models are not interchangeable: language-side hyperparameters must be retuned whenever it is exchanged, as one transferred configuration collapses reproducibly to $20.5$ F1 ($\pm0.8$).

Several additional limitations should be considered when interpreting these results.
First, the \ac{ve} is frozen, so our findings concern adaptation rather than representation learning, and, given its substantial compute cost, the hyperparameter search was conducted only for Primus-1B on CT-RATE, so other configurations are lower bounds and comparisons against the development model are confounded with tuning effort.
Second, because both the automated labeler and its underlying \ac{llm}-extracted training labels are prone to error, clinical macro F1 measures fidelity to a flawed teacher rather than true clinical ground truth. 
Nevertheless, as these errors are clinically plausible \cite{stock2026evaluating}, an F1 score derived from extracted labels yields a substantially more clinically meaningful evaluation than rigid lexical metrics.
Third, at around $50\%$ precision and recall these systems compare design choices and are not deployable report generators.

Overall, our results show that substantial improvements can be obtained through careful choices of input preprocessing, vision encoder, and projector. Although hardware is expected to continue increasing computational capacity, trends toward higher-resolution imaging, higher-resolution vision encoders, and larger language models will continue to increase the demand for efficient vision-token utilization. We therefore expect vision-token allocation to remain a central design consideration for scalable 3D \ac{rrg}.

\section{Acknowledgments}
This work is supported by the Helmholtz Association Initiative and Networking Fund on the HAICORE@JSC partition.

\bibliographystyle{IEEEtran}
\bibliography{references}

\end{document}